\documentclass[letterpaper]{article} % DO NOT CHANGE THIS
\usepackage{aaai2026}  % DO NOT CHANGE THIS
\usepackage{times}  % DO NOT CHANGE THIS
\usepackage{helvet}  % DO NOT CHANGE THIS
\usepackage{courier}  % DO NOT CHANGE THIS
\usepackage[hyphens]{url}  % DO NOT CHANGE THIS
\usepackage{graphicx} % DO NOT CHANGE THIS
\usepackage{natbib}  % DO NOT CHANGE THIS AND DO NOT ADD ANY OPTIONS TO IT
\usepackage{caption} % DO NOT CHANGE THIS AND DO NOT ADD ANY OPTIONS TO IT
\usepackage{algorithm}
\usepackage{algorithmic}
\usepackage{booktabs}
\usepackage{multirow}
\usepackage{xcolor}
\usepackage{lipsum}
\usepackage{subcaption}
\usepackage{caption}
\usepackage{amsfonts}
\usepackage{amsmath}
\usepackage{amssymb}
\usepackage{times}
\usepackage{helvet}
\usepackage{courier}
\usepackage{xcolor}

\usepackage{newfloat}
\usepackage{listings}
\DeclareCaptionStyle{ruled}{labelfont=normalfont,labelsep=colon,strut=off} % DO NOT CHANGE THIS
\floatstyle{ruled}
\newfloat{listing}{tb}{lst}{}
\floatname{listing}{Listing}
\title{% Better Matching, Less Forgetting: A Quality-Guided Matcher for Incremental DETR 
Better Matching, Less Forgetting: A Quality-Guided Matcher for Transformer-based Incremental Object Detection
}
\author {
    Qirui Wu,\textsuperscript{\rm 1}
    Shizhou Zhang,\textsuperscript{\rm 1}\thanks{Shizhou Zhang and De Cheng are co-corresponding authors. 
    This work is done during Qirui Wu's internship at Hikrobot.}
    De Cheng,\textsuperscript{\rm2, \rm 1*}
    Yinghui Xing,\textsuperscript{\rm1}
    Lingyan Ran,\textsuperscript{\rm1}
    Dahu Shi,\textsuperscript{\rm3, \rm4}
    Peng Wang\textsuperscript{\rm1}
}
\affiliations {
    \textsuperscript{\rm 1}ASGO, School of Computer Science, Northwestern Polytechnical University, China\\
    \textsuperscript{\rm 2}School of Telecommunications Engineering, Xidian University, China\\
    \textsuperscript{\rm 3}CCAI, Zhejiang University, China\\
    \textsuperscript{\rm 4}Hikrobot Co., Ltd.\\
    wuqirui@mail.nwpu.edu.cn, szzhang@nwpu.edu.cn, dcheng@xidian.edu.cn, xyh\_7491@nwpu.edu.cn, lran@nwpu.edu.cn, shidahu@zju.edu.cn, peng.wang@nwpu.edu.cn
}

\begin{document}

\maketitle

\begin{abstract}
Incremental Object Detection (IOD) aims to continuously learn new object classes without forgetting previously learned ones. A persistent challenge is catastrophic forgetting, primarily attributed to background shift in conventional detectors. While pseudo-labeling mitigates this in dense detectors, we identify a novel, distinct source of forgetting specific to DETR-like architectures: background foregrounding. This arises from the exhaustiveness constraint of the Hungarian matcher, which forcibly assigns every ground truth target to one prediction, even when predictions primarily cover background regions (i.e., low IoU). This erroneous supervision compels the model to misclassify background features as specific foreground classes, disrupting learned representations and accelerating forgetting. 
To address this, we propose a Quality-guided Min-Cost Max-Flow (Q-MCMF) matcher. To avoid forced assignments, Q-MCMF builds a flow graph and prunes implausible matches based on geometric quality. It then optimizes for the final matching that minimizes cost and maximizes valid assignments. This strategy eliminates harmful supervision from background foregrounding while maximizing foreground learning signals. Extensive experiments on the COCO dataset under various incremental settings demonstrate that our method consistently outperforms existing state-of-the-art approaches. 

% The code is available at https://github.com/fanrena/Q-MCMF.
\end{abstract}

\begin{links}
    \link{Code}{https://github.com/fanrena/Q-MCMF}
\end{links}

\section{Introduction}

Incremental Object Detection (IOD) aims to enable models to learn new object classes continuously while preserving performance on previously learned classes. A critical challenge in IOD is catastrophic forgetting, where conventional detectors (e.g., Faster R-CNN~\cite{fasterrcnn}, DETR~\cite{detr}) tend to lose previously acquired knowledge. Although significant efforts have been devoted to mitigating catastrophic forgetting in object detectors, it remains a persistent and challenging problem in computer vision.

\begin{figure}[t]  % 双栏布局中跨栏放置，优先置顶
    \centering
    % 子图(a)：新类别
    \begin{subfigure}[b]{\linewidth}
        \centering
        % 图片宽度建议略小于textwidth，保留留白
        \includegraphics[width=\linewidth]{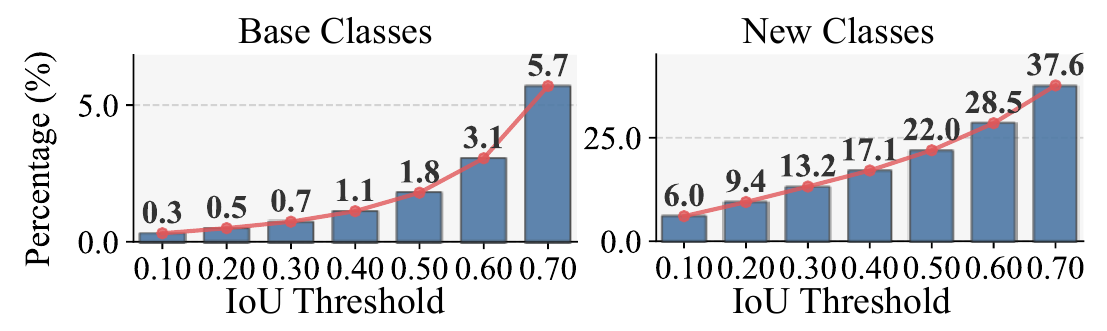}  % 规范文件名（避免空格）
        \caption{Proportion of Matches with IoU $<$ Threshold}  % 明确子图内容
        \label{subfig:new_classes}
    \end{subfigure}
    
    \vspace{8pt}  % 适度缩减间距，避免冗余
    
    % 子图(b)：基础类别
    \begin{subfigure}[b]{\linewidth}
        \centering
        \includegraphics[width=\linewidth]{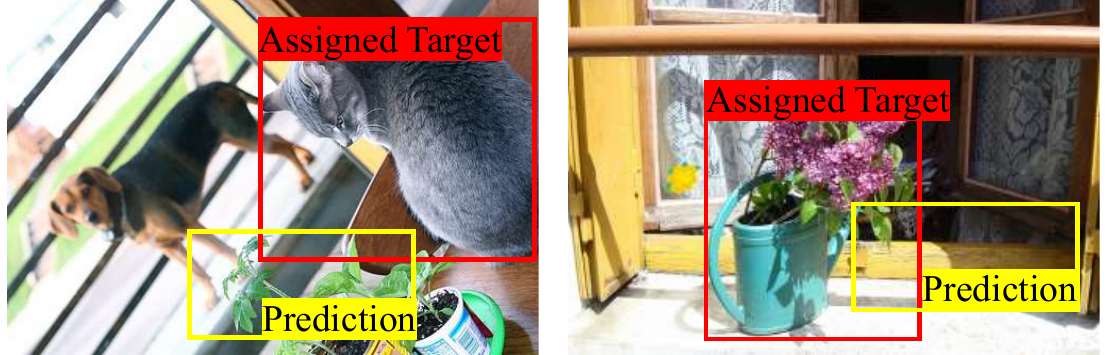}  % 区分文件名
        \caption{Examples of Background Foregrounding}  % 保持标题结构一致
        \label{subfig:base_classes}
    \end{subfigure}
    
    % 整体标题：概括对比关系
    \caption{Quantitative evidence and examples of background foregrounding:(a) shows the proportion of matches with IoU $<$ IoU threshold of Base/New classes at the 40th epoch of 70-10 second phase. 
    (b) shows two examples of background foregrounding.}
    \label{fig:corner_img}
\end{figure}

Previous research identifies background shift~\cite{backgroundshift} as a major source of forgetting. This occurs when objects belonging to previously learned classes (e.g., ``airplane") appear in the training data of a new task but are incorrectly labeled as ``background". 
Optimizing the detector using these mislabeled foreground objects as background not only disrupt the learned foreground representations but also corrupt the background distribution, leading to significantly more severe catastrophic forgetting than inter-class confusion among foreground objects.
Pseudo-labeling has emerged as a simple yet effective strategy to address background shift, applicable to detectors employing both dense predictions~\cite{bpf, repre} 
% (e.g., Faster R-CNN~\cite{bpf, repre}) 
and sparse predictions~\cite{cldetr, dyqdetr}.% (e.g., DETR~\cite{cldetr, dyqdetr}).

Unlike dense detectors, DETR~\cite{detr} distinguishes itself by performing end-to-end object detection without requiring post-processing such as Non-Maximum Suppression (NMS). This capability primarily stems from its use of sparse object queries and the Hungarian matching algorithm. Crucially, the Hungarian matcher serves as the label assignment mechanism, enforcing a strict one-to-one correspondence between predictions and ground truth objects. 
% \textcolor{red}{A key characteristic of this matcher is that it exhaustively assigns all ground truth targets, even if there are no adequate predictions for this target. This inherent property introduces a novel source of forgetting specific to DETR-like architectures, which we term background foregrounding.}
A key characteristic of this matcher is that it exhaustively assigns each target to one prediction, even if the prediction and the target are not geometrically plausible. This exhaustive assignment property introduces a novel source of forgetting specific to DETR-like architectures, which we define as \textit{background foregrounding}.

Background foregrounding arises when a prediction, primarily covering background regions, is forcibly matched to a target due to the exhaustiveness constraint of the Hungarian matcher (see Fig.~\ref{fig:corner_img} (b)). Consequently, the model is incorrectly supervised to classify background features as belonging to a specific foreground class. Similar to background shift, optimizing the model with this erroneous supervision disrupts the learned feature distributions for both the actual foreground classes and the background class, significantly contributing to catastrophic forgetting. Fig.~\ref{fig:corner_img} (a) shows the proportion of matches with IoU below the threshold. At 40th epoch, 5.7\% (old) and 37.6\% (new) of matches have IoU $<$ 0.7, indicating substantial background been assigned as foreground, degrading the detector’s plasticity and stability.

To address this unique challenge inherent in DETR-based IOD, we propose a \textbf{Q}uality-guided \textbf{M}in-\textbf{C}ost \textbf{M}ax-\textbf{F}low (Q-MCMF) matcher.
First, we construct a graph between predictions and ground truth objects with costs, flows, and quality (e.g. IoU) as the property of each edge, then prune all edges with IoU below a threshold to eliminate geometrically implausible matches. This yields a sparse graph where connections represent only valid foreground-background relationships. On this refined graph, we solve the matching problem with dual objectives:
1) Minimizing the total assignment cost to prioritize low-cost matches.
2) Maximizing the number of matched pairs to preserve critical supervision signals.
Crucially, Q-MCMF does not force low-IoU matches, allowing background regions and foreground targets to stay unmatched. This eliminates erroneous supervision that drives background foregrounding. Simultaneously, the matcher maximizes valid matches to prevent slow convergence and performance drops due to sparse supervision after pruning, balancing quantity and quality for robust incremental supervision.
Experiments on the COCO dataset in various settings shows the effectiveness of our method.

Our contributions can be summarized as follows:
\begin{itemize}
    \item To the best of our knowledge, we are the first to identify that background foregrounding as a critical problem that causes catastrophic forgetting in incremental DETR.
    \item To address the background foregrounding problem, we propose a generic Quality guided MCMF Matcher that eliminates low-quality matching while establishing more positive matches.
    \item The proposed approach outperforms existing methods across a variety of single-step and multi-step settings, highlighting its significant effectiveness.
\end{itemize}

\section{Related Works}

\begin{figure*}[!t]
    \centering
    \includegraphics[width=1.0\linewidth]{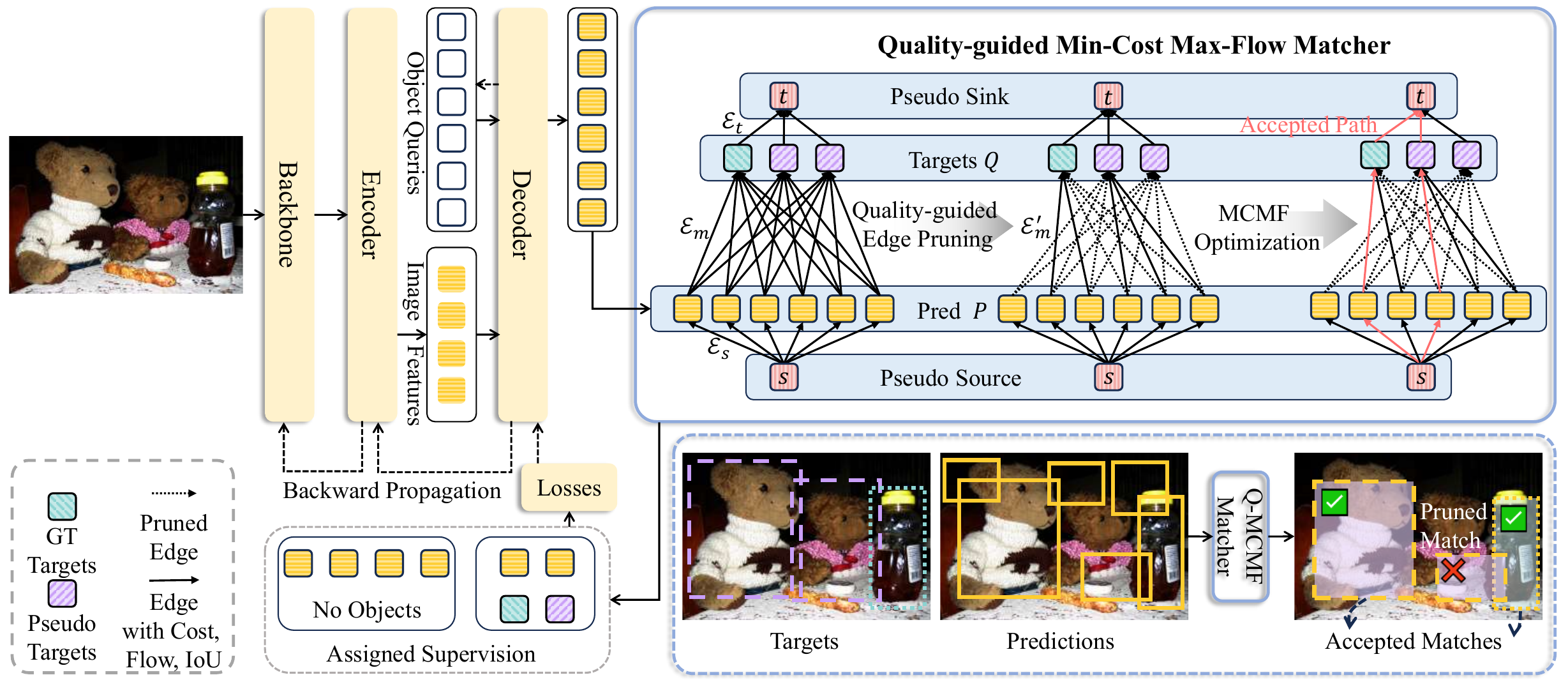}
    \caption{Overall training pipeline of Deformable DETR with Q-MCMF matcher.}
    \label{fig:main_framework}
\end{figure*}

\subsection{Detection Transformers}

DETR~\cite{detr} pioneered end-to-end object detection by eliminating hand-crafted components such as NMS through its use of sparse object queries and a one-to-one Hungarian matcher. While this paradigm simplifies the detection pipeline, DETR encounters significant challenges: the implicit semantics of its object queries~\cite{conditionaldetr, dabdetr, xing2024ms} and slow training convergence~\cite{ddetr, dndetr, groupdetr, hdetr, deim}. 
Addressing the convergence issue specifically, sparse positive supervision has been identified as a primary bottleneck. 
These strategies~\cite{groupdetr,hdetr,deim} demonstrate substantial improvements in accelerating convergence speed and boosting  performance.

While existing methods effectively increase the quantity of supervision, we contend that for incremental learning scenarios, the quality of positive matches is equally critical.

\subsection{Incremental Object Detection}

Incremental Object Detection (IOD) faces the dual challenge of preserving previously acquired detection capabilities while assimilating novel object classes. This complexity stems from two primary vulnerabilities: classification/localization forgetting~\cite{repre, dca} and the background shift phenomenon~\cite{abr, bpf, rgriod, sddgr}.
Background shift~\cite{backgroundshift} occurs when instances of previously learned classes in new-task training data are mislabeled as background, progressively corrupting established representations. Common mitigation strategies include pseudo-labeling~\cite{bpf, cldetr, gda} and exemplar replay~\cite{cldetr, abr, sddgr, rgriod, repre}, which aim to preserve knowledge of prior classes. 

Diverging from these approaches, our work investigates a more fundamental aspect: the role of label assignment mechanisms within DETR-based IOD frameworks, where we identify critical limitations requiring novel solutions.

\subsection{Label Assignment}

Defining positive/negative training samples is fundamental in object detection, as the quality of assignment critically impacts performance. 
Current assignment strategies operate under predefined correspondence rules, broadly categorized into two paradigms:
1. One-to-many assigners (e.g., Max-IoU, ATSS~\cite{atss}, OTA~\cite{ota}, SimOTA~\cite{yolox}) allocate multiple predictions per ground truth.
2. One-to-one assigners enforce strict pairwise matching, exemplified by DETR's~\cite{detr} Hungarian algorithm, enabling NMS-free end-to-end detection.

Our analysis identifies a flaw in the Hungarian matcher: its exhaustive matching forces low-quality predictions to targets, causing background foregrounding. This disrupts feature distributions catastrophically during incremental learning via erroneous supervision. Hence, we propose Q-MCMF Matcher to resolve this problem.

\section{Method}

\subsection{Preliminary}

\textbf{IOD Formulation.} In IOD, model training is structured across \(n\) sequential learning stages, where each stage introduces a novel set of classes for detection. 
Formally, let \(\mathcal{T} = \{\mathcal{T}_1, \mathcal{T}_2, \ldots, \mathcal{T}_n\}\) represent the complete class set incrementally acquired by detector \(\mathcal{F}\), with \(\mathcal{T}_i \cap \mathcal{T}_j = \emptyset\) for all \(i \neq j\). 
The training dataset for stage \(k\) is denoted as \(\mathcal{D}_k = \{\mathcal{X}_{k}, \mathcal{Y}_{k}\}\), where \(\mathcal{X}_{k}\) contains input images and \(\mathcal{Y}_{k}\) provides corresponding annotations. 
A critical characteristic of IOD is that while images in \(\mathcal{X}_{k}\) may contain objects from any class in \(\mathcal{T}\), only those belonging to \(\mathcal{T}_k\) are explicitly annotated. 
The primary challenge lies in updating the detector from \(\mathcal{F}_{k-1}\) to \(\mathcal{F}_k\) using only the current dataset \(\mathcal{D}_k\) (i.e., without access to previous data \(\{\mathcal{D}_1, \dots, \mathcal{D}_{k-1}\}\)), so as to improve its capability on \(\mathcal{T}_k\) while avoiding catastrophic forgetting on all previous classes \(\bigcup_{i=1}^{k-1}\mathcal{T}_i\).

\textbf{Detection Transformer.}
Our work focuses on the DETR architecture, specifically Deformable DETR~\cite{ddetr}. Deformable DETR is a fully end-to-end object detector comprising four key components: 1) a backbone that extracts initial multi-scale feature maps from the input image; 2) a transformer encoder that encodes these features; 3) object queries, a set of learnable embeddings encoding positional and semantic priors, that interact with the encoded features to locate objects; and 4) a transformer decoder that uses the object queries to attend to relevant regions within the encoder's output feature map. Through iterative cross-attention layers, the decoder refines each query into an object representation. 
These final decoded representations are fed into prediction heads for bounding box regression and category classification.

The source of DETR's end-to-end characteristic lies in its prediction-target matcher. Predictions are assigned to ground truth targets via the Hungarian Matcher, which finds a strict one-to-one, bipartite assignment between predictions and targets by minimizing global cost. This matching strategy eliminates the need for NMS post-processing, enabling DETR's end-to-end nature. 

However, the exhaustive matching property of the Hungarian algorithm—which compels every ground truth target to be assigned a prediction, even when no geometrically compatible prediction exists—induces knowledge erosion. This occurs through misclassification of background as foreground, disrupting the model's stability-plasticity balance. 
% Next, we will elaborate on the Hungarian matcher and introduce necessary concepts for our method.

\textbf{Hungarian Matcher.} 
In DETR, the Hungarian matcher serves as the label assigner to assign labels to predictions in a strict one-to-one correspondence. Specifically, let the predictions be $P=\{p_i \mid i\in \mathbb{N}, 1 \leq i \leq N_p \}$ where $N_p$ is the number of predictions, and the targets be $Q=\{q_i \mid i\in \mathbb{N}, 1\leq i \leq N_q \}$ where $N_q$ is the number of targets. There is a complete bipartite graph $\mathcal{G}_m=(P, Q; \mathcal{E}_m)$ where $\mathcal{E}_m=\{(p_i, q_j) \mid \forall p_i \in P,\forall q_j\in Q\}$ is the collection of edges in the graph with a non-negative cost $c(p_i, q_j)$, usually to be the loss between $p_i$ and $q_j$. The ${\bf C}=\{c(p_i, q_j)\mid \forall p_i \in P,\forall q_j\in Q\}$ is also known as the cost matrix. The goal of the Hungarian matcher is to find an assignment of the targets to the predictions ${\cal M}\subset \mathcal{E}_m$ while the total cost of assignment is minimum. That is to say, the optimization goal of the Hungarian matcher is
\begin{equation}
    \begin{aligned}
    &\mathop{\arg\min}\limits_{\cal M} \sum_{(p_i, q_j)\in \mathcal{M}} c(p_i, q_j), \\ 
    & \text{s.t.} \quad
    \begin{cases}
        \sum_{j=1}^{N_p} \mathbb{I}[(p_i,q_j) \in \mathcal{M}] \leq 1,  \forall p_i \in P \\[0.2em]
        \sum_{i=1}^{N_q} \mathbb{I}[(p_i,q_j) \in \mathcal{M}] \leq 1,  \forall q_j \in Q \\[0.2em]
        |{\cal M}| = \min(N_p,N_q),
    \end{cases}
    \end{aligned}
    \label{eq:hungarian_restriction}
\end{equation}
where $\mathbb{I}(\cdot)$ is the indicator function. Here, the first two restrictions limit edges in $\cal M$ from sharing vertex, ensuring a strict one-to-one assignment. The last constraint dictates that the cardinality of \(\cal M\) equals the smaller of the cardinalities of $P$ and $Q$. This ensures that every single node in $P$ or $Q$ is exhaustively assigned to a target/prediction.

The fundamental limitation lies in the complete bipartite graph structure and its requirement for perfect matching on one node set as shown in Eq.~\ref{eq:hungarian_restriction}. The Hungarian matcher's exhaustive assignment induces background foregrounding, which catastrophically corrupts feature distributions in incremental learning scenarios. To address this, we propose a simple yet effective Quality-guided Min-cost Max-flow (Q-MCMF) Matcher.

\subsection{Quality-guided Min-Cost Max-Flow Matcher}

To mitigate the background foregrounding problem, our matching framework should satisfy four key properties:
1) Strict one-to-one correspondence in assignments.
2) Minimized matching cost over the selected pairs.
3) Selective exclusion capability that allows targets without adequate matches to be excluded from the assignment, thereby eliminating background foregrounding.
4) Maximized matched pairs, as sparse matching hampers model from convergence
and limits performance as suggested in prior works~\cite{groupdetr, hdetr, deim}.
To meet these requirements, we reformulate the matching task as a min-cost max-flow (MCMF) problem. 
This framework provides flexible cost matrix manipulation while simultaneously satisfying all four requirements through a unified optimization process.

\textbf{Graph Construction.} To start, we transform the complete bipartite graph $\mathcal{G}_m$ into a directed flow network $\mathcal{G} = (s,t, P, Q; \mathcal{E})$, where $s$ and $t$ are pseudo source and pseudo sink respectively. As shown in Fig.~\ref{fig:main_framework}, the directed edge set $\mathcal{E}$ is defined as the union of three edge sets
$\mathcal{E} = \mathcal{E}_s \cup \mathcal{E}_{m} \cup \mathcal{E}_t$ where
\begin{equation}
\begin{aligned}
\begin{cases}
\mathcal{E}_s = \{(s, p_i) \mid p_i \in P\}\\
\mathcal{E}_m = \{(p_i, q_j) \mid p_i \in P, q_j \in Q\}\\
\mathcal{E}_t = \{(q_j, t) \mid q_j \in Q\}.
\end{cases}
\end{aligned}
\end{equation}
Every edge has three attributes: cost $c(p_i, q_j)$, flow $f(p_i, q_j)$, and IoU $\phi(p_i, q_j)$ between $p_i$ and $q_j$.  For every $p_i \in P$ and $q_j \in Q$, we assign $c(s, p_i) = 0$, $c(q_j, t) = 0$, $\phi(s, p_i) = 1$, and $\phi(q_j, t) = 1$ to ensure zero contribution of $\mathcal{E}_s$ and $\mathcal{E}_t$ to the results. 
The flow capacity is set to $f(e) = 1$ for every edge $e \in \mathcal{E}$ to ensure one-to-one assignment.

\textbf{Quality-guided Edge Pruning.} With $P$ and $Q$ densely connected by $\mathcal{E}_{m}$, solving the min-cost max-flow problem on $\mathcal{G}$ deteriorates to a matching problem on $\mathcal{G}_m$, and it only fosters the minimization of cost. To reduce background foregrounding, we introduce quality-guided edge pruning to prune edges with low IoU. Specifically, we refine $\mathcal{E}_{m}$ by
\begin{equation}
\begin{aligned}
    \mathcal{E}_{m}^\prime = 
        \left\{ (p_i, q_j^{old}) \mid \phi(p_i, q_j^{old}) \geq \alpha \right\} \cup \\
        \left\{ (p_i, q_j^{new}) \mid \phi(p_i, q_j^{new}) \geq \beta \right\},
\end{aligned}
\end{equation}
where $q_j^{old}$ and $q_j^{new}$ represent target labels sourced from pseudo-labels (learned tasks) or ground truth (current task), respectively.  $\alpha$ and $\beta$ is the different threshold for different targets. % Accompanied with our slover, this allows some targets with no adequate match to be dismissed from the final assignment.

The rationale for employing distinct thresholds for different targets stems from the disparate predictive capabilities of the model with respect to seen and unseen classes. 
For seen classes, accurate predictions and effective assignment are more readily achievable. Conversely, for unseen classes, achieving high-quality assignment proves challenging due to the model's lack of prior exposure. 
A high threshold can reduce the number of positive samples and slow down the convergence rate, whereas a low threshold, although it increases the positive matches, intensifies the background foregrounding issue. 
Consequently, the parameters $\alpha$ and $\beta$ govern the balance between stability and plasticity within our method.

\begin{table}[t]
\centering

\setlength{\tabcolsep}{3pt}
\begin{tabular}{l|l|cccccc}
\toprule
\textbf{(1)} & \textbf{Methods} & \textbf{AP} & \textbf{AP\textsubscript{50}} & \textbf{AP\textsubscript{75}} & \textbf{AP\textsubscript{S}} & \textbf{AP\textsubscript{M}} & \textbf{AP\textsubscript{L}} \\
\midrule
\multirow{11}{*}{\rotatebox{90}{40-40}}
 & LwF*        & 17.2 & 25.4 & 18.6 & 7.9  & 18.4 & 24.3 \\
 & CL-DETR*   & 42.0 & 60.1 & 45.9 & 24.0 & 45.3 & 55.6 \\
 & SDDGR*     & \underline{43.0} & \underline{62.1} & \underline{47.1} & \underline{24.9} & \underline{46.9} & \underline{57.0} \\
 & DyQ-DETR*  & 42.4 & 60.5 & 45.9 & 23.9 & 46.3 & 56.7 \\
 & RILOD      & 29.9 & 45.0 & 32.0 & 15.8 & 33.0 & 40.5 \\
 & SID        & 34.0 & 51.4 & 36.3 & 18.4 & 38.4 & 44.9 \\
 & ERD        & 36.9 & 54.5 & 39.6 & 21.3 & 40.4 & 47.5 \\
 & CL-DETR    & 39.2 & 56.1 & 42.6 & 21.0 & 42.8 & 52.6 \\
 & DyQ-DETR   & 41.4 & 59.7 & 44.9 & 24.1 & 45.2 & 54.3 \\
 & DCA        & 42.8 & 58.4 & --   & --   & --   & --   \\
 & Ours       & \textbf{44.3} & \textbf{59.5} & \textbf{49.2} & \textbf{27.1} & \textbf{49.9} & \textbf{59.8} \\
\midrule
\multirow{11}{*}{\rotatebox{90}{70-10}} 
 & LwF*        & 7.1  & 12.4 & 7.0  & 4.8  & 9.5  & 10.0 \\
 & CL-DETR*   & 40.4 & 58.0 & 43.9 & 23.8 & 43.6 & 53.5 \\
 & SDDGR*     & 40.9 & 59.5 & 44.8 & 23.9 & 44.7 & 54.0 \\
 & DyQ-DETR*  & \underline{42.4} & \underline{60.4} & \underline{46.3} & \underline{24.5} & \underline{45.7} & \underline{57.5} \\
 & RILOD      & 24.5 & 37.9 & 25.7 & 14.2 & 27.4 & 33.5 \\
 & SID        & 32.8 & 49.0 & 35.0 & 17.1 & 36.9 & 44.5 \\
 & ERD        & 34.9 & 51.9 & 37.4 & 18.7 & 38.8 & 45.5 \\
 & CL-DETR    & 35.8 & 53.5 & 39.5 & 19.4 & 41.5 & 46.1 \\
 & DyQ-DETR   & 39.5 & 56.4 & 43.1 & 22.5 & 43.1 & 53.0 \\
 & DCA        & 41.3 & 59.2 & --   & --   & --   & --   \\
 & Ours       & \textbf{43.4} & \textbf{62.0} & \textbf{47.6} & \textbf{28.2} & \textbf{46.9} & \textbf{56.0} \\
\bottomrule
\end{tabular}
\caption{IOD results (\%) on COCO 2017 under the 40-40 and 70-10 setting using protocol (1). * indicates results with exemplar replay. The best performance in each is presented in bold, and the second best is presented with underline.}
\label{tab:main_results}
\end{table}

\begin{table}[t]
\centering

\setlength{\tabcolsep}{3pt}
\begin{tabular}{l|l|cccccc}
\toprule
\textbf{(2)} & \textbf{Methods} & \textbf{AP} & \textbf{AP\textsubscript{50}} & \textbf{AP\textsubscript{75}} & \textbf{AP\textsubscript{S}} & \textbf{AP\textsubscript{M}} & \textbf{AP\textsubscript{L}} \\
\midrule
\multirow{8}{*}{\rotatebox{90}{40-40}}
 & LwF*       & 23.9 & 41.5 & 25.0 & 12.0 & 26.4 & 33.0 \\
 & CL-DETR*   & 37.5 & 55.1 & 40.3 & 20.9 & 40.8 & 50.7 \\
 & DyQ-DETR*  & \underline{39.7} & \underline{57.5} & \underline{43.0} & \underline{21.6} & \underline{42.9} & \underline{53.8} \\
 & iCaRL      & 33.4 & 52.0 & 36.0 & 18.0 & 36.4 & 45.5 \\
 & ERD        & 36.0 & 55.2 & 38.7 & 19.5 & 38.7 & 49.0 \\
 & CL-DETR    & 36.2 & 52.6 & 39.5 & 18.7 & 39.5 & 49.4 \\
 & DyQ-DETR   & 39.1 & 57.1 & 42.5 & 21.3 & 42.7 & 51.8 \\
 & Ours       & \textbf{42.4} & \textbf{60.5} & \textbf{46.4} & \textbf{25.5} & \textbf{46.5} & \textbf{55.2} \\
\midrule
\multirow{8}{*}{\rotatebox{90}{70-10}} 
 & LwF*       & 24.5 & 36.6 & 26.7 & 12.4 & 28.2 & 35.2 \\
 & CL-DETR*   & 40.1 & 57.8 & 43.7 & 23.2 & 43.2 & 52.1 \\
 & DyQ-DETR*  & \underline{41.9} & \underline{60.1} & \underline{45.8} & \underline{24.1} & \underline{45.3} & \underline{55.8} \\
 & iCaRL      & 35.9 & 52.5 & 39.2 & 19.1 & 39.4 & 48.6 \\
 & ERD        & 36.9 & 55.7 & 40.1 & 21.4 & 39.6 & 48.7 \\
 & CL-DETR    & 34.0 & 48.0 & 37.2 & 15.5 & 37.7 & 49.7 \\
 & DyQ-DETR   & 39.6 & 57.6 & 43.5 & 23.4 & 43.3 & 51.8 \\
 & Ours       & \textbf{43.1} & \textbf{61.3} & \textbf{47.5} & \textbf{27.9} & \textbf{46.7} & \textbf{55.7} \\
\bottomrule
\end{tabular}
\caption{IOD results (\%) on COCO 2017 under the 40-40 and 70-10 setting using protocol (2). * indicates results with exemplar replay. The best performance in each is presented in bold, and the second best is presented with underline.}
\label{tab:no_overlap_results}
\end{table}

\textbf{MCMF Optimization.} The objective is to find a set of paths \(\mathcal{P} = \{s \to p_i \to q_j \to t \mid p_i \in P, q_j \in Q\}\) that simultaneously minimizes total cost and maximizes flow. Following quality-guided edge pruning, we formulate the MCMF problem's objectives as:  
\begin{equation}
\begin{aligned}
    &\mathop{\arg\min}\limits_{\mathcal{P}} \sum_{(p_i, q_j)\in \mathcal{P}} c(p_i, q_j), \\
    & \text{s.t.} 
    \begin{cases}
        \sum_{j=1}^{N_p} \mathbb{I}[(p_i,q_j) \in \mathcal{P}] \leq 1, \forall p_i \in P \\[0.2em]
        \sum_{i=1}^{N_q} \mathbb{I}[(p_i,q_j) \in \mathcal{P}] \leq 1, \forall q_j \in Q \\[0.2em]
        |\mathcal{P}| = \mathop{\arg\max}\limits_{|\mathcal{P}|}\left(\sum_{(s, p_i,q_j,t)\in \mathcal{P}}f(s, p_i, q_j, t)\right). \\
    \end{cases}
\end{aligned}
\end{equation}
Here, akin to Eq.~\ref{eq:hungarian_restriction}, the first two restrictions ensure that paths in $\mathcal{P}$ do not share intermediate edges and guarantee strict one-to-one matching. The last constraint limits the flow of path set $\mathcal{P}$ to the maximum flow in graph $\mathcal{G}$. This not only ensures the number of maximum matching pairs but also allows implausible matches to be discarded in the final assignment, thereby preventing the background foregrounding problem.
The solution reduces to the intermediate edges $\mathcal{M}^\prime = \{ (p_i, q_j)\mid \forall(s, p_i, q_j, t)\in \mathcal{P} \}$, which constitute the final matching assignment.

This optimization naturally yields an assignment of the prediction target $\mathcal{M}^\prime$ that satisfies our four design criteria and effectively mitigates the background foregrounding problem.

\subsection{Training Objectives}

After obtaining the prediction-target assignment $\mathcal{M}^\prime$, we compute the same loss function as that used in conventional Deformable DETR. 
Here, we denote $p^c_i$ and $p^b_i$ as the class prediction and its bounding box prediction of $p_i$ while $q^c_j$ and $q^b_j$ as the class label and its bounding box annotation of $q_i$. The loss function for each matched pair $(p_i,q_j) \in \mathcal{M}^\prime$ consists of
\begin{equation}
\begin{aligned}
    \mathcal{L}_{fg} = & \sum_{(p_i, q_j)\in \mathcal{M}^\prime}(\lambda_{\text{focal}}\mathcal{L}_{\text{focal}}(p^c_i, q^c_j) + \\
    & \lambda_{L1}\mathcal{L}_{L1}(p^b_i, q^b_j) + \lambda_{\text{giou}}\mathcal{L}_{\text{giou}}(p^b_i, q^b_j)),
\end{aligned}
    \label{eq:loss}
\end{equation}
where $\mathcal{L}_{\text{focal}}$ denotes the focal loss for predicted class $p^c_i$ and target class $q^c_j$, $\mathcal{L}_{L1}$ and $\mathcal{L}_{\text{giou}}$ represent bounding box regression losses between predicted box $p^b_i$ and target box $q^b_j$, and $\lambda$ terms are the loss coefficients to balance the trade-off between different loss components in the optimization objective. Predictions not assigned to any target ($P_{u} = \{p_i \mid p_i \notin \mathcal{M}^\prime \}$) are supervised as background class. The classification loss for these predictions is the standard focal loss formulated as
\begin{equation}
    \mathcal{L}_{\text{bg}} = \sum_{p_i \in P_{u}} \mathcal{L}_{\text{focal}}(p^c_i, \varnothing),
\end{equation}
where $\varnothing$ represents background.
Finally, the overall loss function is computed as 
\begin{equation}
    \mathcal{L} = \mathcal{L}_{fg} + \lambda_{\text{bg}}\mathcal{L}_{\text{bg}}.
\end{equation}

\section{Experiments}

\begin{table}[t]
\centering

\setlength{\tabcolsep}{0.5pt}
\begin{tabular}{lcccccc}
\toprule
\multirow{2}{*}{\textbf{Methods}} & \multicolumn{4}{c}{\textbf{$40\mbox{-}10\times4$}} & \multicolumn{2}{c}{\textbf{$40\mbox{-}20\times2$}} \\
\cmidrule(lr){2-5} \cmidrule(lr){6-7}
%  & \textbf{+40–50} & \textbf{+50–60} & \textbf{+60–70} & \textbf{+70–80} & \textbf{+40–60} & \textbf{+60–80} \\
& {\small\textbf{+40–50}} & {\small\textbf{+50–60}} & {\small\textbf{+60–70}} & {\small\textbf{+70–80}} & {\small\textbf{+40–60}} & {\small\textbf{+60–80}} \\
\midrule
CF         & 5.8  & 5.7  & 6.3  & 3.3  & 10.7 & 9.4  \\
RILOD      & 25.4 & 11.2 & 10.5 & 8.4  & 27.8 & 15.8 \\
SID        & 34.6 & 24.1 & 14.6 & 12.6 & 34.0 & 23.8 \\
ERD        & 36.4 & 30.8 & 26.2 & 20.7 & 36.7 & 32.4 \\
CL-DETR*   & --   & --   & --   & 28.1 & --   & 35.3 \\
ACF        & 39.1 & 35.4 & 32.0 & 30.3 & 39.3 & 36.6 \\
SDDGR*     & 42.3 & 40.6 & 40.0 & 36.8 & 42.5 & 41.1 \\
DCA        & \textbf{44.0} & \underline{41.1} & \underline{39.2} & \underline{37.2} & \underline{42.7} & \underline{40.3} \\
Ours       & \underline{43.2} & \textbf{41.2} & \textbf{40.0} & \textbf{38.0} & \textbf{44.3} & \textbf{42.3} \\
\bottomrule
\end{tabular}
\caption{IOD results (\%) on COCO 2017 under the $40\mbox{-}10\times 4$ and $40\mbox{-}20\times 2$ setting using protocol (1). * indicates the results are obtained with exemplar replay. The best performance in each is presented in bold, and the second best is presented with underline.}
\label{tab:multi_stage_results}
\end{table}

\subsection{Experimental Setup}

\textbf{Dataset and Evaluation Metrics.}
For fair comparison, we follow the training protocols as in previous works~\cite{cldetr, ilod} and evaluate our method on the widely adopted COCO 2017 dataset, which consists of 80 different classes in natural scenes. 
The AP, AP\textsubscript{50}, AP\textsubscript{75}, AP\textsubscript{S}, AP\textsubscript{M}, and AP\textsubscript{L} of standard COCO metrics are used for performance evaluation.

\textbf{Protocols.}
We evaluated our methods with two different protocols. (1) We follow previous works~\cite{ilod} and split dataset with classes. In this protocol, images for training are overlapped for different tasks while their annotations are not. (2) Following ~\cite{cldetr}, we split the dataset into non-overlapping splits for training. In this protocol, both classes and unseen data are introduced sequentially and learned incrementally, which is a more practical simulation to the real-world application.  Therefore, we evaluate our method with protocol (2) unless specified. We evaluate our method with two-phase and multiple-phase settings. For two-phase setting($A\mbox{-}B$), we evaluate with $70\mbox{-}10$ setting and $40\mbox{-}40$ setting, where $A$ are the number of base classes and $B$ represents new classes. 
For multiple-phase setting($C\mbox{-}D\times E$), we evaluate our method with $40\mbox{-}10\times4$ and $40\mbox{-}20\times2$ settings, where $C$ denotes the number of base classes, $D$ denotes the number of subsequent new classes, and $E$ denotes the number of new phases.

\textbf{Implementation Details.} Following ~\cite{cldetr}, we build our method on top of the Deformable DETR without iterative bounding box refinement and the two-stage variant. To address background shift problem, we adopt DKD proposed in CL-DETR. 
The backbone we adopted is ResNet-50~\cite{resnet} pre-trained on ImageNet~\cite{imagenet}. We train our detector with AdamW~\cite{adamw} for 50 epoch. We set our quality guided edge pruning hyperparameter to $\alpha=0.7$ and $\beta=0.5$, which is determined through a grid search. We also implemented our method on DN-DETR~\cite{dndetr} and DAB-DETR~\cite{dabdetr} with the same hyperparameter as we used in Deformable DETR without further grid search for optimal hyperparameters.
% Training configurations for the initial stage are consistent with CL-DETR. 
More implementation details can be found in the appendix.

\subsection{Comparison with SOTA Methods}

\textbf{Two-Phase Settings.}
% We compare our method with methods incorporates exemplar reaply method such as RILOD*, SID*, ERD*, CL-DETR*, SDDGR* and DyQ-DETR* and exemplar free method such as LwF, CL-DETR, DyQ-DETR, and previous SOTA method DCA. Note that our method do not use exemplar replay. Tab.~\ref{tab:main_results} presents results under protocol (1). It shows that, in two-phase settings, our proposed method consistently outperforms the aforementioned methods under different protocols with significant margins, even methods replays 10\% of the entire dataset. Tab.~\ref{tab:no_overlap_results} presents results under protocol (2). It shows that our method surpasses DyQ-DETR by 3.5\% in 70-10 setting while achieves +3.3\% gain in 40-40 setting. It also surpass DyQ-DETR* by 1.2\% and 2.7\% in 70-10 and 40-40 respectively. 
We compare our method with existing approaches, including exemplar replay-based methods (LwF*~\cite{lwf}, CL-DETR*~\cite{cldetr}, SDDGR*~\cite{sddgr}, DyQ-DETR*~\cite{dyqdetr}) and exemplar-free methods (RILOD~\cite{rilod}, SID~\cite{sid}, ERD~\cite{erd}, CL-DETR~\cite{cldetr}, DyQ-DETR~\cite{dyqdetr}), along with the previous state-of-the-art method DCA~\cite{dca}. Notably, our approach operates without exemplar replay. 
As presented in Tab.~\ref{tab:main_results} under protocol (1), our method consistently outperforms all comparison methods across different incremental learning settings by significant margins. Our method surpass its baseline, CL-DETR, by 5.1\% and 7.6\% in 40-40 and 70-10 settings in AP metric. It also surpasses previous best results by 1.0\% and 1.3\%  in 70-10 and 40-40 respectively. It is worth noting that previous best results replay generative data (SDDGR*) or 10\% labeled data from the dataset (DyQ-DETR*) while ours preserve knowledge in an example-free manner.

Further results under protocol (2) (Tab.~\ref{tab:no_overlap_results}) demonstrate that our method surpasses DyQ-DETR by 3.5\% in the 70-10 setting and achieves a 3.3\% gain in the 40-40 setting. When compared to DyQ-DETR* with exemplar replay, our method also shows superior performance with gains of 1.2\% (70-10) and 2.7\% (40-40). The consistent performance gains across both protocols validate our method's superiority to different task configurations. It also highlights the background foregrounding problem as a critical problem in incremental transformer detector.

\begin{figure}[!t]
    \centering
    \includegraphics[width=1.0\linewidth]{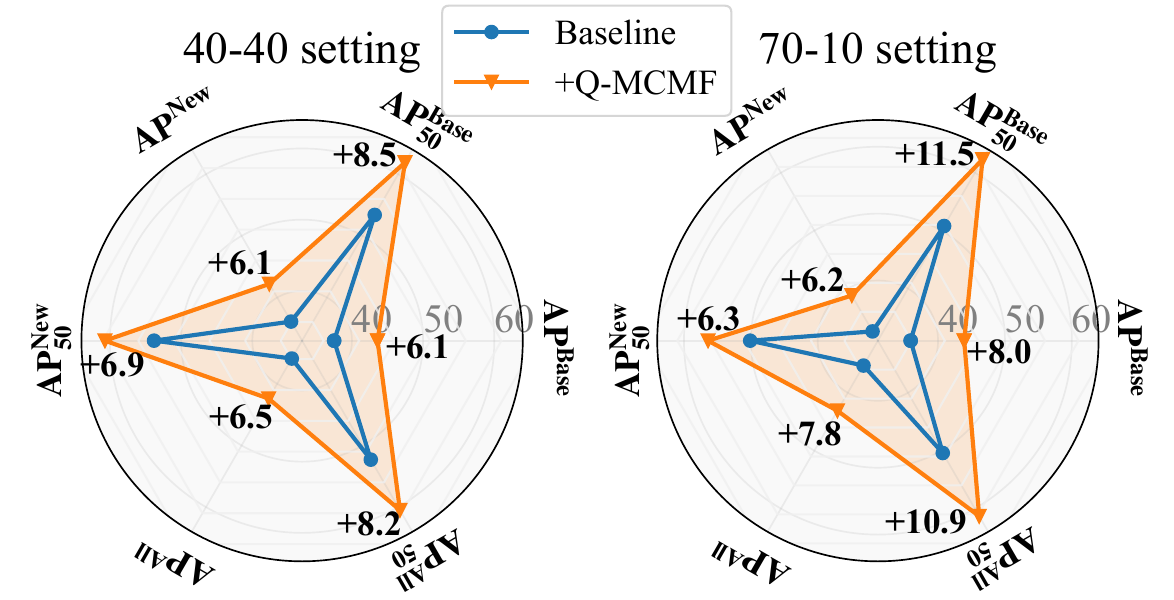}
    \caption{Effectiveness analysis of Q-MCMF Matcher.}
    \label{fig:qmcmf}
\end{figure}

\textbf{Multiple-Phase Settings.}
% In Tab.~\ref{tab:multi_stage_results}, we present results under a more demanding setting, where the model need to learn incrementally in multiple phases. For fair comparison, we use the same data split as previous works for fair comparison. Notably, our method outperforms exemplar based methods CL-DETR and SDDGR by 9.9\% and 1.2\% in $40\mbox{-}10\times 4$. Our method also outperforms previous SOTA method DCA by 0.8\% in $40\mbox{-}10\times 4$ setting. In $40\mbox{-}20\times 2$, our method surpass exemplar based methods CL-DETR and SDDGR by 7.0\% and 1.2\% while surpassing DCA by 2.0\%.  
In Tab.~\ref{tab:multi_stage_results}, we report the results under the more demanding multi-phase incremental learning setting, using the same data splits as prior works for fair comparison. 
Notably, in the $40\mbox{-}10\times 4$ configuration, our method outperforms exemplar-based methods CL-DETR and SDDGR by 9.9\% and 1.2\% respectively, while surpassing the previous state-of-the-art DCA by 0.8\%. Similarly, in the $40\mbox{-}20\times 2$ setting, our approach achieves significant gains of 7.0\% over CL-DETR, 1.2\% over SDDGR, and 2.0\% over DCA. These consistent improvements across different phase configurations demonstrate our method's robustness in handling complex multi-phase learning scenarios. This also shows that solving background foregrounding problem can achieve consistent performance gain.

\subsection{Ablation Study}

\begin{figure}[!t]
    \centering
    \includegraphics[width=1.0\linewidth]{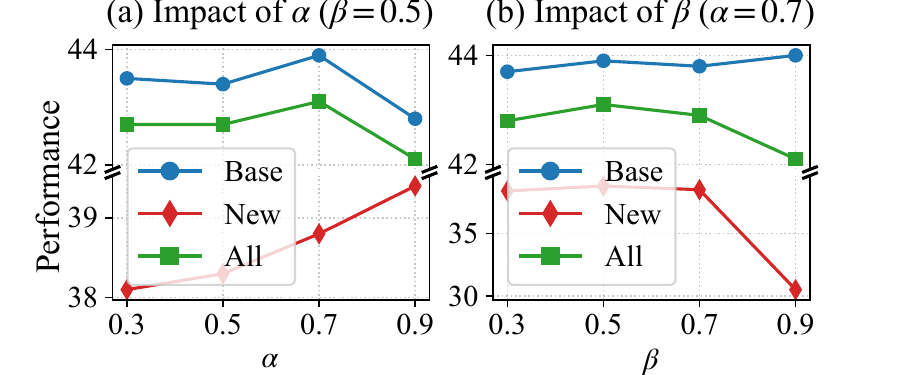}
    \caption{Stability-Plasticity Balance with different $\alpha$ and $\beta$.}
    \label{fig:hyperparam_results}
\end{figure}

\textbf{Effectiveness Analysis of Q-MCMF Matcher.} 
In Fig.~\ref{fig:qmcmf}, we present a comprehensive ablation study of our proposed Q-MCMF matcher, comparing its performance against the baseline. Our baseline implementation is CL-DETR without exemplar replay.

Regarding stability (Base classes), Q-MCMF demonstrates substantial improvements: +6.1\% AP in the 40-40 setting and +8.0\% AP in the 70-10 setting, indicating superior retention of prior knowledge. For plasticity (New classes), it achieves gains of +6.1\% AP (40-40) and +6.2\% AP (70-10), confirming effective adaptation to novel categories. The overall performance (All classes) shows consistent enhancement with improvements of +6.5\% AP (40-40) and +7.8\% AP (70-10). These gains, complemented by significant increases in AP\textsubscript{50}, validate Q-MCMF's ability to mitigate background foregrounding while optimizing the stability-plasticity balance in incremental object detection.

\textbf{Stability-Plasticity Balance.} 
Fig.~\ref{fig:hyperparam_results} presents the performance of base and new classes under the 70-10 incremental setting for different $\alpha$ and $\beta$ values. 
% To study their individual effects, one parameter was fixed while varying the other, with baseline values set at $\alpha=0.7$ and $\beta=0.5$ when investigating each parameter's impact. 
To study their individual effects, one parameter was varied while the other was fixed: $\beta$ was fixed at 0.5 for $\alpha$ studies, and $\alpha$ was fixed at 0.7 for $\beta$ studies.
As shown in Fig.~\ref{fig:hyperparam_results}, optimal stability for base classes is observed at $\alpha=0.7$ in (a) while optimal plasticity for new classes is at $\beta=0.5$ in (b), with deviations in either direction causing performance deterioration. 
% This indicates the $\alpha$ parameter primarily governs model stability and the $\beta$ primarily influence model plasticity.
% Higher $\alpha$/$\beta$ values significantly reduce matches, leading to an 1.1\% performance drop in base classes and 8.9\% drop in new classes. 
% This occurs because excessive threshouds restricts positive matches for classes, resulting in insufficient training, slower convergence, and ultimately reduced performance. 
By observing new performance in Fig.~\ref{fig:hyperparam_results} (a), new class performance consistently improves with increasing $\alpha$. This occurs because elevated $\alpha$ reduces base class matching, thereby limiting interference from base to new class feature distributions. Symmetrically, in Fig.~\ref{fig:hyperparam_results} (b), base class performance improves with higher $\beta$ due to its equivalent protective effect against interference.
The All curve shows that the overall performance represents a trade-off between stability and plasticity. This indicates the balance of stability and plasticity is manipulated by both $\alpha$ and $\beta$ instead of $\alpha$ solely controls stability and $\beta$ solely controls plasticity.

\begin{figure}[!t]
    \centering
    \includegraphics[width=1.0\linewidth]{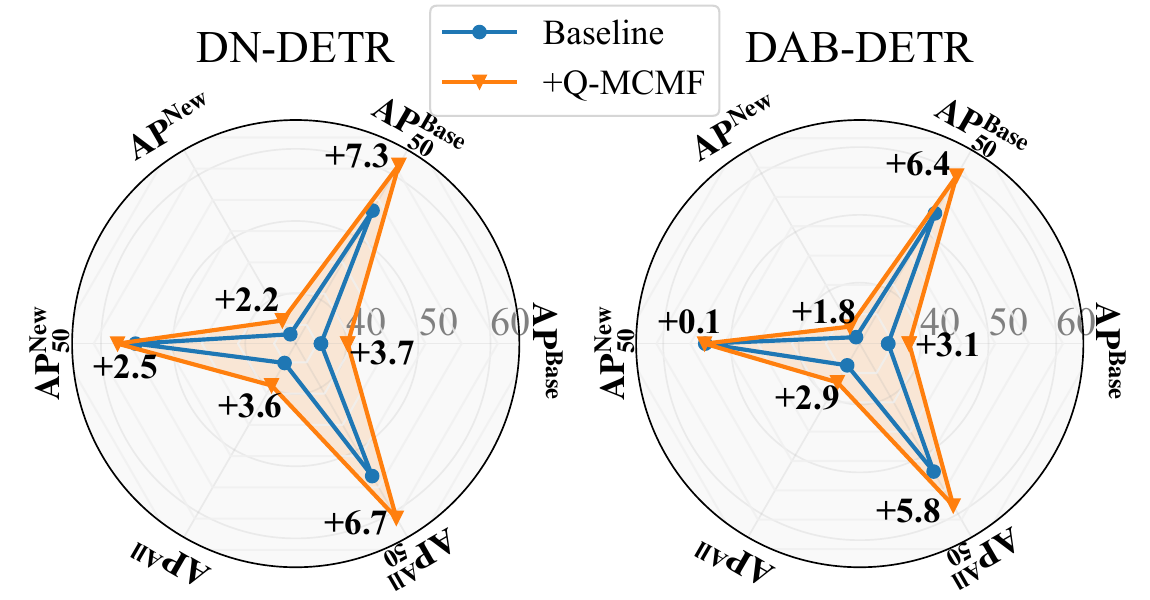}
    \caption{Effectiveness of Q-MCMF Matcher on DN-DETR and DAB-DETR under 70-10 setting.}
    \label{fig:qmcmf_dn}
\end{figure}

\textbf{Effectiveness of Q-MCMF on Different Architectures.} To validate the robustness of our proposed method, we further evaluate our method on two extra architecture: DN-DETR and DAB-DETR. As shown in Fig.~\ref{fig:qmcmf_dn}, we present the results of these two detectors under 70-10 setting. Specifically, our method demonstrates consistent improvements across both architectures. For DN-DETR, our method obtain a gain of 3.7\% compared to our baseline while AP\textsubscript{50} achieves a substantial +7.3\% gain. New-class performance also increases, with AP increasing by +2.2\% and AP\textsubscript{50} by 2.5\%. Consequently, overall AP is increased by +3.6\% to 39.7\%, and AP\textsubscript{50} is increased by +6.7\% to 60.9\%.
Similarly, DAB-DETR also exhibits noticeable gains. 
% These results validate our approach’s efficacy in enhancing incremental detection across diverse DETR frameworks. 
These consistent improvements across both DN-DETR and DAB-DETR architectures collectively demonstrate the robustness and generalizability of our approach, validating its effectiveness for transformer-based incremental object detection.

\begin{figure}[!t]
    \centering
    \includegraphics[width=1.0\linewidth]{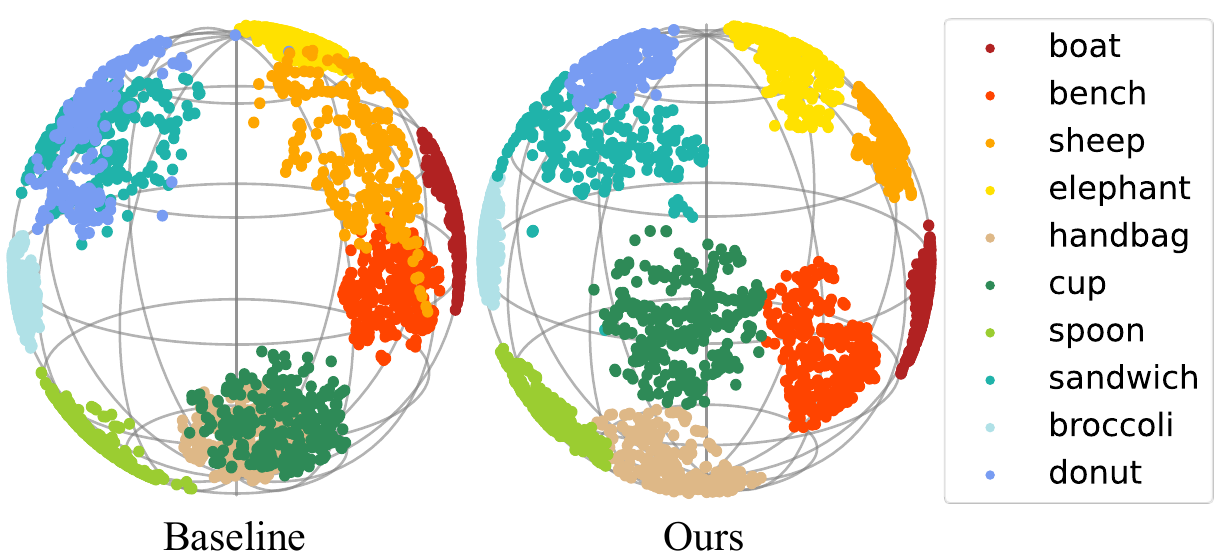}
    \caption{t-SNE visualization of learned categories' features after the second phase training of the 70-10 setting.}
    \label{fig:tsne}
\end{figure}

\textbf{t-SNE Visualization.} 
To further verify that our method causes less distortion to learned feature semantics compared to baselines, we perform qualitative analysis via t-SNE~\cite{tsne} visualization. 
We random select 10 categories and extract the features of the decoder's last layer and map them to 3D space using t-SNE. As shown in Fig.~\ref{fig:tsne}, features from our method exhibit stronger intra-class compactness (e.g., sheep and donut) and clearer inter-class separability (e.g., sheep and elephant), whereas baselines show more scattered clusters and overlapping between classes. This indicates that our method better preserves the original semantic structure of features, confirming its advantage in mitigating feature distortion and forgetting.

\begin{figure}[!t]
    \centering
    \includegraphics[width=1.0\linewidth]{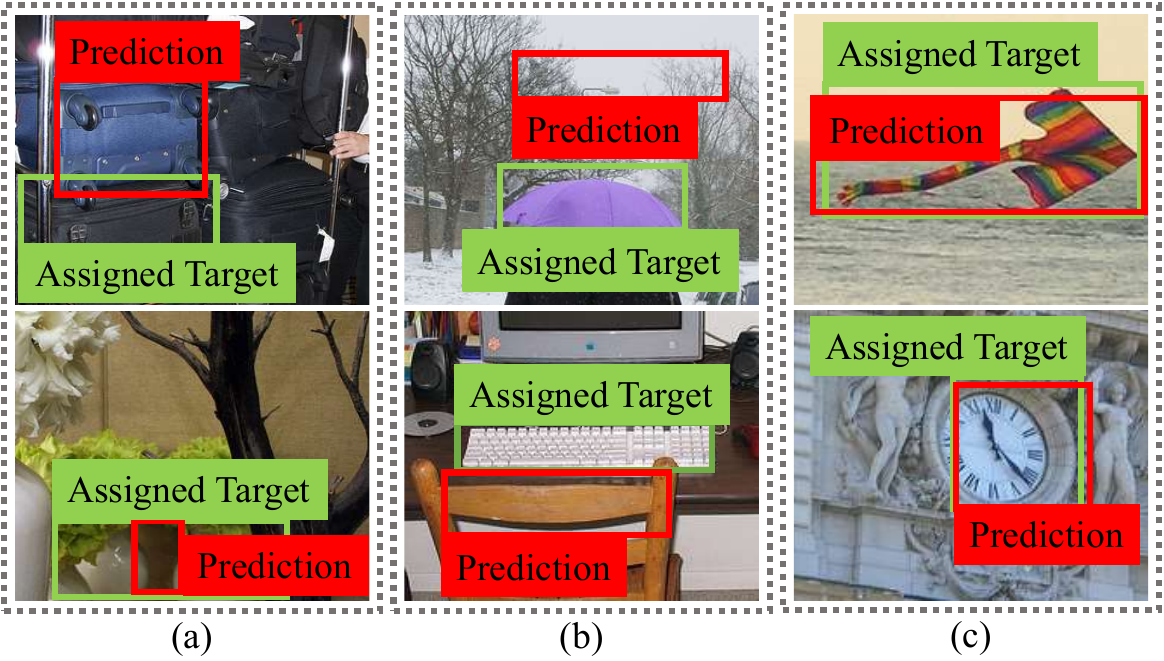}
    \caption{Zoomed in examples of matches. (a) shows examples of rejected match from base classes, (b) shows examples of rejected match from new classes, and (c) shows examples of accepted match.}
    \label{fig:visualizations}
\end{figure}

\textbf{Match Visualization.} Fig.~\ref{fig:visualizations} validates our motivation by contrasting Q-MCMF and Hungarian matcher decisions, showing how Q-MCMF eliminates geometrically implausible matches. Crucially, the Hungarian Matcher persists in assigning targets to geometrically incompatible predictions for both base classes and new classes, causing background foregrounding. In contrast, Q-MCMF explicitly rejects such spatially inconsistent matches through quality-guided edge pruning, which significantly reduces background foregrounding and prevents erroneous supervision during training. More samples are presented in the appendix.

\section{Conclusion}

In transformer-based incremental object detection, we identify background foregrounding as a novel source of catastrophic forgetting. This phenomenon stems from misalignment between background predictions and foreground targets, directly caused by the Hungarian matcher's exhaustive one-to-one matching requirement. To address this limitation, we propose the Q-MCMF matcher, which incorporates quality-guided edge pruning to eliminate geometrically implausible matches while simultaneously optimizing for minimal matching cost and maximal valid assignments. By preventing erroneous supervision from background foregrounding, our method effectively mitigates forgetting. Our work highlights the criticality of architecture-specific forgetting sources in IOD, with Q-MCMF offering a simple yet effective solution. We hope that our research will offer fundamental insights into forgetting mitigation for DETR-based incremental detectors, facilitating progress in this area.

\section*{Acknowledgements}

This work was supported in part by the National Natural Science Foundation of China (NSFC) under Grant 62576282, 62176198, 62476223, 62576262,62476226; in part by the National Key Research and Development Program of China under Grant 2024YFF1306501; in part by Innovation Capability Support Program of Shaanxi (Program No. 2024ZC-KJXX-043, 2024GX-YBXM-135); in part by Natural Science Basic Research Program of Shaanxi Province (2024JC-DXWT-07).

\bibliography{aaai2026}

\end{document}